\documentclass[conference]{IEEEtran}
\IEEEoverridecommandlockouts

\usepackage{cite}
\usepackage{amsmath,amssymb,amsfonts}

\usepackage{graphicx}
\usepackage{textcomp}
\usepackage{algorithm}
\usepackage{algpseudocode}
\usepackage{svg}
\usepackage{multirow}
\usepackage{hyperref}
\usepackage{amsmath}
\usepackage{blindtext}
\usepackage{xcolor}
\usepackage{float}
\usepackage{subcaption}

\def\BibTeX{{\rm B\kern-.05em{\sc i\kern-.025em b}\kern-.08em
    T\kern-.1667em\lower.7ex\hbox{E}\kern-.125emX}}
\begin{document}

\title{An Atmospheric Correction Integrated LULC Segmentation Model for High-Resolution Satellite Imagery
}
  
\author{
\IEEEauthorblockN{
Soham Mukherjee$^{a,b}$, Yash Dixit$^{a}$, Naman Srivastava$^{a}$, 
Joel D Joy$^{a}$, \\ Rohan Olikara$^{a}$, Koesha Sinha$^{a}$, 
Swarup E$^{a}$, Rakshit Ramesh$^{a}$}
\IEEEauthorblockA{
$^{a}$\textit{IUDX Programme Unit, Indian Institute of Science, Bengaluru}\\
}
\IEEEauthorblockA{
$^{b}$\textit{Département de Biologie, Chimie et Géographie, Université du Québec à Rimouski, Rimouski} \\
}
}

\maketitle

\begin{abstract}
The integration of fine-scale multispectral imagery with deep learning models has revolutionized land use and land cover (LULC) classification. However, the atmospheric effects present in Top-of-Atmosphere sensor measured Digital Number values must be corrected to retrieve accurate Bottom-of-Atmosphere surface reflectance for reliable analysis. This study employs look-up-table-based radiative transfer simulations to estimate the atmospheric path reflectance and transmittance for atmospherically correcting high-resolution CARTOSAT-3 Multispectral (MX) imagery for several Indian cities. The corrected surface reflectance data were subsequently used in supervised and semi-supervised segmentation models, demonstrating stability in multi-class (buildings, roads, trees and water bodies) LULC segmentation accuracy, particularly in scenarios with sparsely labelled data.
\end{abstract}

\begin{IEEEkeywords}
Multispectral Remote Sensing, Atmospheric Correction, Radiative transfer, Image Segmentation, Semi-supervised Learning, Land Use Land Cover.
\end{IEEEkeywords}

\section{Introduction}
\label{sec:intro}

The advent of fine-scale space-borne multispectral acquisition has improved the field of both terrestrial and aquatic remote sensing (RS), encompassing from water quality detection \cite{Ibrahim2018,Blix2018} to spectral indices calculation, land use and land cover (LULC) classification etc. \cite{vali2020deep}. An efficient LULC classification can be key in environmental monitoring, urban planning, and resource management. Current land use labels for urban planning, such as buildings, roads, and water bodies, rely on limited sources like Java Open Street Maps or model predictions from Microsoft and Google. Previous approaches to LULC prediction have tackled issues like class imbalance and specific terrains, implementing supervised and semi-supervised approaches \cite{chen2022semi,dixit2024}. Among such approaches, the U-Net architecture, primarily designed for biomedical image segmentation, has become highly effective in land use classification tasks 
Similarly, DeepLab v3+, an advanced deep learning model by Google DeepMind, has significantly enhanced the approach to land use classification \cite{fan2022land}. Besides these, Cross Pseudo Supervision (CPS) based learning with a consistency regularization approach containing two segmentation networks has been used for LULC segmentation on sparsely labelled images, proving its effectiveness over supervised models \cite{dixit2024,chen2021semi}. However, raw multispectral images, typically captured as Top-of-Atmosphere (TOA) Digital Number (DN) values may not be directly used in deep learning-based approaches; the pixel values are affected by multiple scatterings and spectral absorption in the atmosphere, e.g. interaction(s) with aerosols, water vapour, and ozone, which suppresses the signal from the Bottom-of-Atmosphere (BOA) target surface \cite{Singh2014}. The spectral values used for each pixel in deep learning-based LULC classification approaches should correspond accurately to the surface characteristics for a robust prediction, hence converting TOA DN values to Bottom-of-Atmosphere (BOA) reflectance through atmospheric correction (AC) to obtain analysis-ready data (ARD) is vital.\\

\begin{figure}[h]
    \centering
    \includegraphics[scale=0.3]{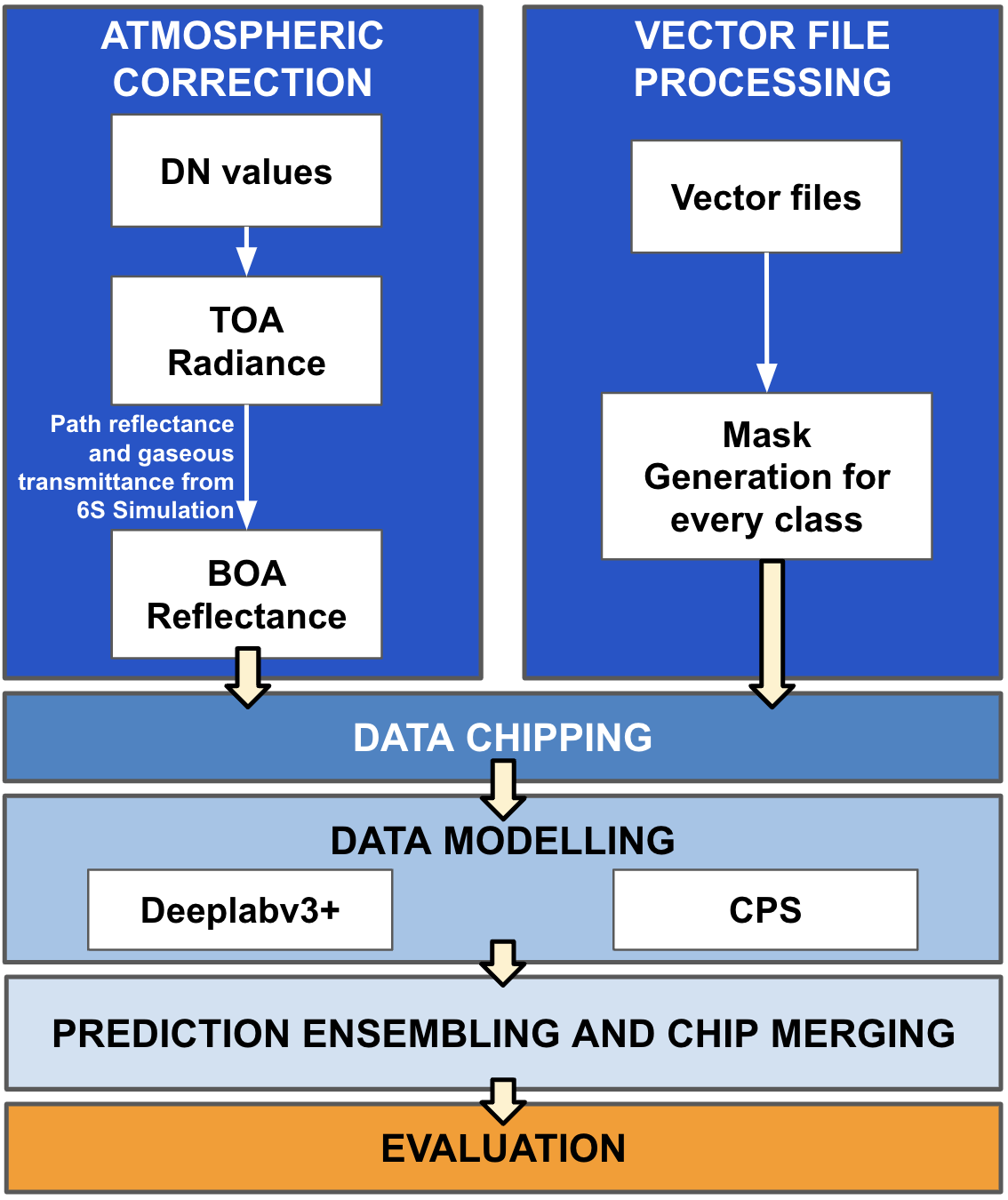}
    \caption{Workflow for LULC segmentation}
    \label{fig:ard-1}
\end{figure}

Several empirical, e.g., Dark Object Subtraction (DOS) \cite{isprs_dos_2013} or physics-based radiative transfer simulations \cite{Singh2014} for atmospherically correcting multispectral images have been developed over time. The current study adopts the Second Simulation of the Satellite Signal in the Solar Spectrum (6S, \cite{6s}) radiative transfer model to mitigate atmospheric effects by adjusting for the atmospheric transmittance and path reflectance to retrieve robust estimates of spectral surface reflectance. This approach was tested on CARTOSAT-3 MX; a very fine-scale multispectral (MX) sensor in the VIS-NIR domain. The AC obtained surface reflectance was then input to Semi-supervised and supervised segmentation models to assess the sensor performance in multi-class segmentation for sparsely labelled data.
\begin{figure*}[!htb]
    \centering
    \includegraphics[scale=0.42]{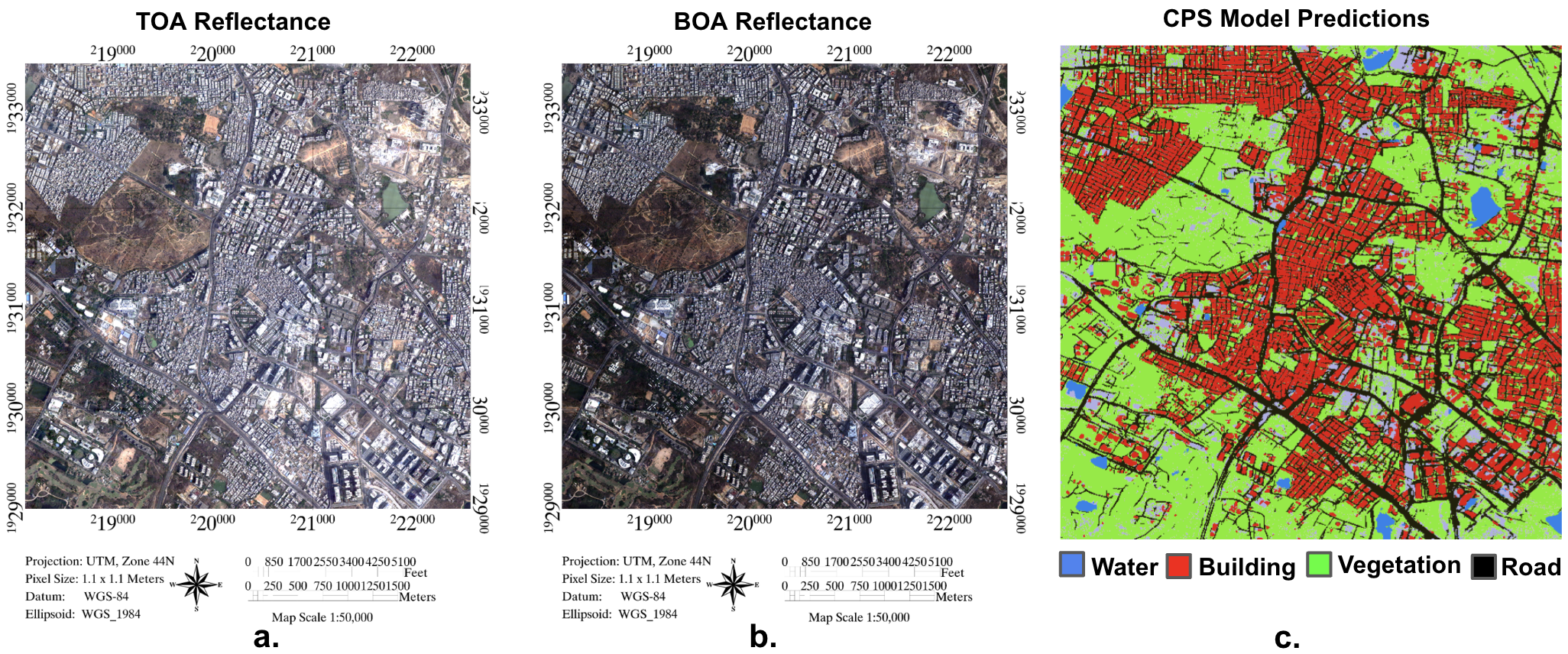}
    
    \label{fig:ard}
\end{figure*}

\begin{figure*}[!htb]
    \centering
    \includegraphics[scale=0.4]{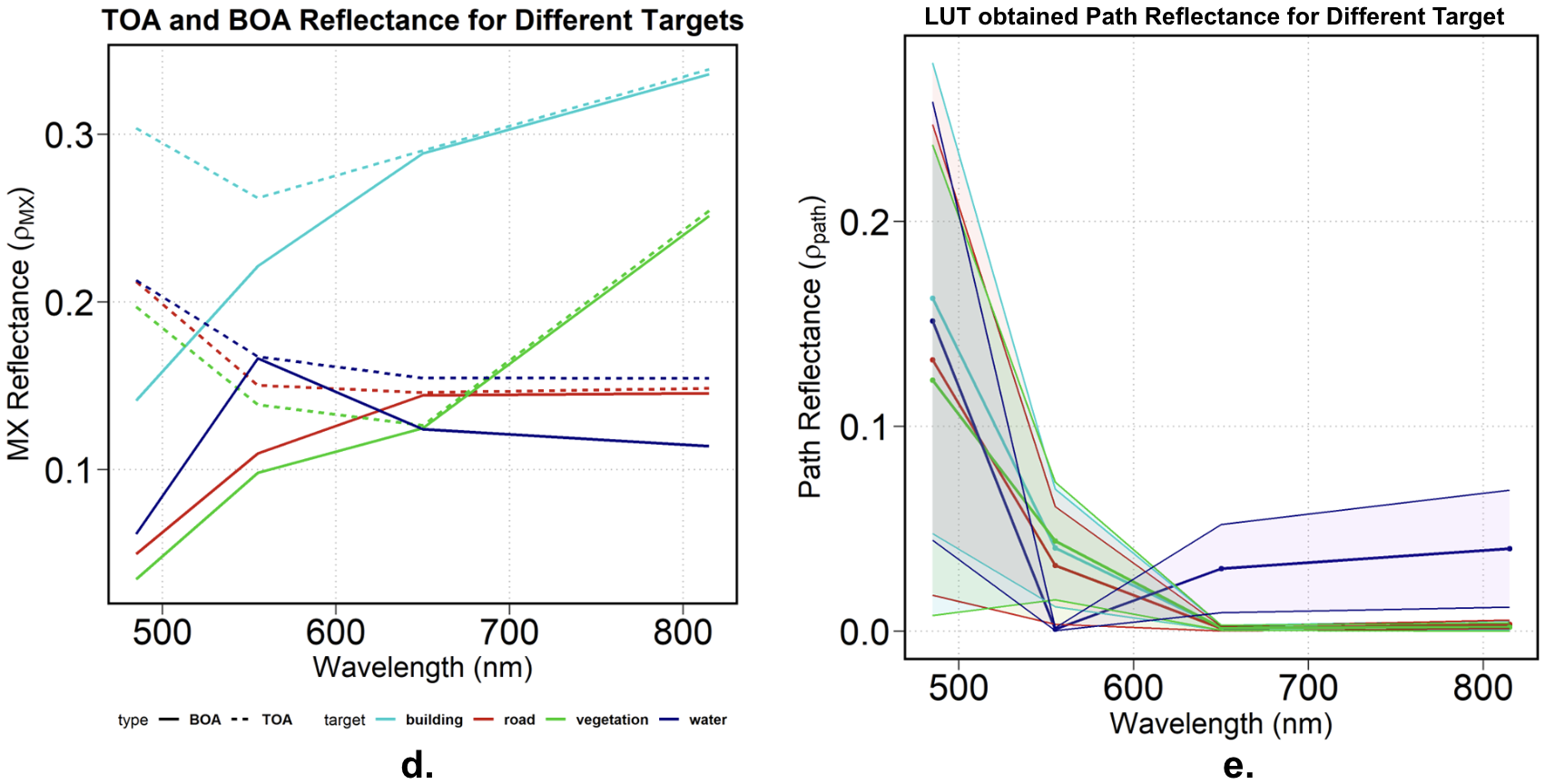}
    \caption{(\textbf{a}) TOA reflectance for the subset MX image acquired over Hyderabad, (\textbf{b}) AC obtained BOA reflectance for the same extent as panel (\textbf{a}),(\textbf{c}) CPS model predictions, (\textbf{d}) inter-comparison of TOA and BOA reflectance for four LULC classes, (\textbf{e}) LUT obtained mean Path reflectance for the LULC classes. The shaded regions show the 1$\sigma$ uncertainty for each class. }
    \label{fig:ard}
\end{figure*}

\section{Materials and Methods}
\label{sec:m_m}
\subsection{Materials}
\label{sssec:materials}
\subsubsection{Raster Data}
In this study, the very high-resolution (around $1.134m ^ {2}$/px) multispectral (Blue: 0.45-0.52 $\mu$m, Green: 0.52-0.59 $\mu$m, Red: 0.62-0.68 $\mu$m, NIR: 0.77-0.86 $\mu$m) scenes were used acquired over several Indian cities ($n=6$) by the MX sensor mounted on Indian Space Research Organisation's (ISRO) CartoSAT-3 satellite. Each tile was chipped to a subset of the image with sufficient LULC class heterogeneity as a preprocessing step as mentioned in Fig.\ref{fig:ard-1}. The associated ground truth of land use classes was obtained following a binary mask approach rasterizing LULC vectors as detailed in \cite{dixit2024}.
\subsubsection{Vector Data}
Multiple sets of vector-type data were acquired for training and evaluation sets differently. For training, We acquired vector files for different land uses enclosed within the geographical extent of MX images captured over Bangalore, Mumbai, Pune, Varanasi and Delhi, respectively. For our evaluation set, we included data from Google Buildings and Microsoft Footprints for buildings and the JOSM vector for the remaining four classes (Roads, Vegetation, Water Bodies and Others) over a subset of developed region in Hyderabad.
\subsection{Methods}
\label{ssub:methods}
\subsubsection{Atmospheric Correction}
\label{sssec:atm}
The TOA apparent reflectance ($\rho^*$) can be linked to the BOA reflectance under the assumption of Lambertian surface \cite{6s}:

\begin{align}
\label{eq:rtm_atmos}
\rho^*\left(\theta_s, \theta_v, \Delta \phi, \lambda\right)= & \ T_g\left(\theta_s, \theta_v, \lambda\right) \times \bigg[\rho_{r a}\left(\theta_s, \theta_v, \Delta \phi, \lambda\right) \nonumber\\
& + \frac{T\left(\theta_s, \lambda\right) T\left(\theta_v, \lambda\right) \rho_s(\lambda)}{1-S(\lambda) \rho_s(\lambda)}\bigg]
\end{align}
Where, $\theta_s$, $\theta_v$, $\Delta \phi$ and $\lambda$ are the sun zenith angle, viewing zenith angle, relative azimuth angle and wavelength, respectively. $\rho$ denotes apparent reflectance obtained as $\rho = \frac{\pi L}{F_0 \mu_s}$. $L$ is the acquired radiance converted from raw DN value, $F_0$ is the exterrerstrial solar irradiance and $\mu s$ is the cosine of $\theta_s$. $\rho_s$ is the BOA (surface)reflectance, $\rho_{ra}$ is the atmospheric path reflectance from the combined interaction of aerosols and molecules. $T\left(\theta_s, \lambda\right)$ and $T\left(\theta_v, \lambda\right)$ are the upward and downward atmospheric transmittance, $S$ is the spherical albedo and $T_g$ refers to gaseous (Ozone and Water Vapour) transmittance. 

6S \cite{6s} was used to simulate Look-Up Tables(LUTs) for the solution vector for a large set of AC coefficients (obtained by rearraging the unknown components in Eq. \ref{eq:rtm_atmos} apart from the $\rho_s$). The coefficients characterize 1. $\rho_{ra}$, a function of aerosol optical thickness and atmospheric molecular Rayleigh scattering and 2. Gaseous absorption and transmittance, which are functions of atmospheric water vapour and ozone, primarily along with the target surface elevation and acquisition geometry. The LUT was exploited to interpolate the correction coefficients for the estimated gaseous concentration, aerosol optical thickness (AOT), surface elevation and viewing geometry ($\theta_s, \theta_v$) averaged over the extent for each band present in MX scenes. 

\subsubsection{LULC Model Training}
\label{ssec:lulc}
\textbf{Cross Pseudo Supervision}: The Cross Pseudo Supervision (CPS) for this work follows the implementation of \cite{dixit2024} to train the LULC model.  
Once a pair of models are trained using CPS, we implement two post-processing techniques:\\
\textbf{Prediction Ensembling}: We generate predictions using both models, the final prediction for each chip is an average of the softmax probabilities from both models.\\
\textbf{Prediction Merging}: We merge output chips which are present in a sliding window format, based on geospatial coordinates using max pooling to recreate the full image. This method leverages overlap in chips to ensure broader context. 
\subsubsection{LULC Model Evaluation}
Evaluating our model's performance using standard metrics like Accuracy or Dice score is challenging due to sparse ground truth labels. These metrics often misclassified predictions against the sparse labels. Instead, we use Recall (Eq. \ref{eq:recall}), which ensures proper assessment by minimizing False Negatives (FN) and maximizing True Positives (TP).

\begin{equation}
\label{eq:recall}
    Recall = \frac{TP}{TP+FN}
\end{equation}

\begin{table*}[!ht]
    \centering    
    \begin{tabular}{|p{2cm}|c|c|c|c|c|}
    \hline
        \textbf{Model} & \textbf{Input} & \textbf{Trees} & \textbf{Buildings} & \textbf{Water} & \textbf{Roads} \\
        \hline        
         \multirow{2}{*}{CPS} & ARD & 93.114 & 74.3087 & 86.8159 & 62.0745 \\
         \cline{2-6}
          & Raw & 78.5390 & 65.1801 & 66.0376 & 58.4839 \\
         \hline
         \multirow{2}{*}{Deeplabv3+} & ARD & 50.9215 & 57.8633  & 50.3246& 40.0537  \\
         \cline{2-6}
          & Raw & 16.6227 & 53.2109 & 70.0857 & 0.0410 \\
         \hline
    \end{tabular}
    \caption{Recall Score For Supervised and Semi-Supervised Model calculated on a subsection of the Hyderabad ARD image covering $24.6 \mathrm{km}^{2}$}
    \label{tab:recall-scores}
\end{table*}

\section{Results and Discussion}
The LUT-based atmospheric correction (AC) was applied to a subset of an MX image over Hyderabad, captured on March 12, 2023. The AC-corrected image (Fig. \ref{fig:ard}b) showed significant improvement by removing atmospheric effects, which typically cause a hazy appearance in the uncorrected $\rho^{}$ (Fig. \ref{fig:ard}a). Fig. \ref{fig:ard}d demonstrates a reduction in BOA reflectance values inversely proportional to the spectral channels, with the blue band being mostly affected by the ARD procedure. This is likely due to correcting the 'haze' effect caused by Rayleigh scattering at the scale of $\lambda^{-4}$ \cite{Singh2014}, particularly noticeable in the blue band of the MX image. However, in urban scenes, aerosols also significantly contribute to atmospheric scattering. A continental aerosol model, combined with molecular scattering, was used to derive the path reflectance ($\rho_{ra}$) (Fig. \ref{fig:ard}e). Notable spectral variation was observed between terrestrial LULC and water classes, with a sharper drop in $\rho_{ra}$ for the green band, potentially due to the highest signal-to-noise ratio in green for water, reducing atmospheric ambiguity for TOA $\rho^*$ \cite{Pan2022}. It is important to note that our approach currently lacks a calculation for water surface glint \cite{Cox1954}, which could further improve the accuracy of the blue and green bands in the future.
On obtaining the AC corrected ARD image, we applied both supervised and semi-supervised approaches mentioned in \ref{ssec:lulc}. We used prediction ensembling method to combine the two models' outputs within CPS. We then applied a 0.35 threshold to the softmax values of the ensembled output to produce the final binary result, as shown in Fig.\ref{fig:ard}a. After merging the predictions into a single image, recall scores were assessed which are shown in Table \ref{tab:recall-scores}, calculated on a subsection of the Hyderabad Cartosat image covering $24.6km^{2}$. Our CPS approach frequently misclassifies shadows on buildings as the 'Other' class instead of correctly identifying them as part of the building class. This inaccuracy may arise from the similarity in spectral signatures between these two classes in the training data. To address this, the model training phase could incorporate a dynamic weighting strategy in the loss functions to better predict minority classes or create special regions of interest for such areas. 

\section{Conclusion}
We demonstrated that semi-supervised learning models effectively classify LULC classes from ARD derived from fine-scale multispectral imagery. The atmospheric correction process successfully removed atmospheric effects, yielding accurate spectral reflectance. While the method performed well for terrestrial LULC classes, incorporating surface glint correction is needed for better water classification The semi-supervised model, CPS showed demonstrated a significant improvement in the recall scores after the implementation of ARD as compared to the original 11- bit raw image.

\bibliographystyle{unsrt} 
\bibliography{main} 

\end{document}